\documentclass[letterpaper, 10 pt, conference]{ieeeconf}  

\IEEEoverridecommandlockouts                              

\usepackage{graphicx} 
\usepackage{amsmath} 
\usepackage{amssymb}  
\usepackage{pifont} 
\usepackage{booktabs}
\usepackage{paralist}
\usepackage{stfloats}
\usepackage{float}
\usepackage{caption}
\usepackage{cuted}
\usepackage{dsfont} 
\usepackage{subcaption}
\usepackage[hidelinks]{hyperref}
\usepackage{listings}
\usepackage{xcolor}
\lstdefinelanguage{YAML}{
  keywords={true,false,null,y,n},
  keywordstyle=\color{blue},
  basicstyle=\ttfamily\small,
  sensitive=false,
  comment=[l]{\#},
  morecomment=[s]{/*}{*/},
  commentstyle=\color{gray}\ttfamily,
  stringstyle=\color{orange},
  morestring=[b]',
  morestring=[b]",
}

\newcommand{\cmark}{\ding{51}} 
\newcommand{\xmark}{\ding{55}} 
\title{\LARGE \bf
Find the Fruit: Zero-Shot Sim2Real RL for Occlusion-Aware \\Plant Manipulation
}

\author{Nitesh Subedi$^{1*}$, Hsin-Jung Yang$^{1*}$, Devesh K. Jha$^{2}$, Soumik Sarkar$^{1}$
\thanks{$^{1}$Department of Mechanical Engineering, Iowa State University, USA
{\tt\small \{nitesh, hjy, soumiks\}@iastate.edu}}%
\thanks{$^{2}$Mitsubishi Electric Research Laboratories, USA
        {\tt\small devesh.dkj@gmail.com}}%
\thanks{*Equal contribution, \href{https://drive.google.com/file/d/1AIowX6JTGccliKXr55fBmcsXxST3Bwd2/view?usp=sharing}{Appendix: https://tinyurl.com/ftfappendix}
}}

\begin{document}
\maketitle

\thispagestyle{empty}
\pagestyle{empty}

\begin{strip}
\vspace{-6.5em}
    \centering
    \includegraphics[width=\linewidth]{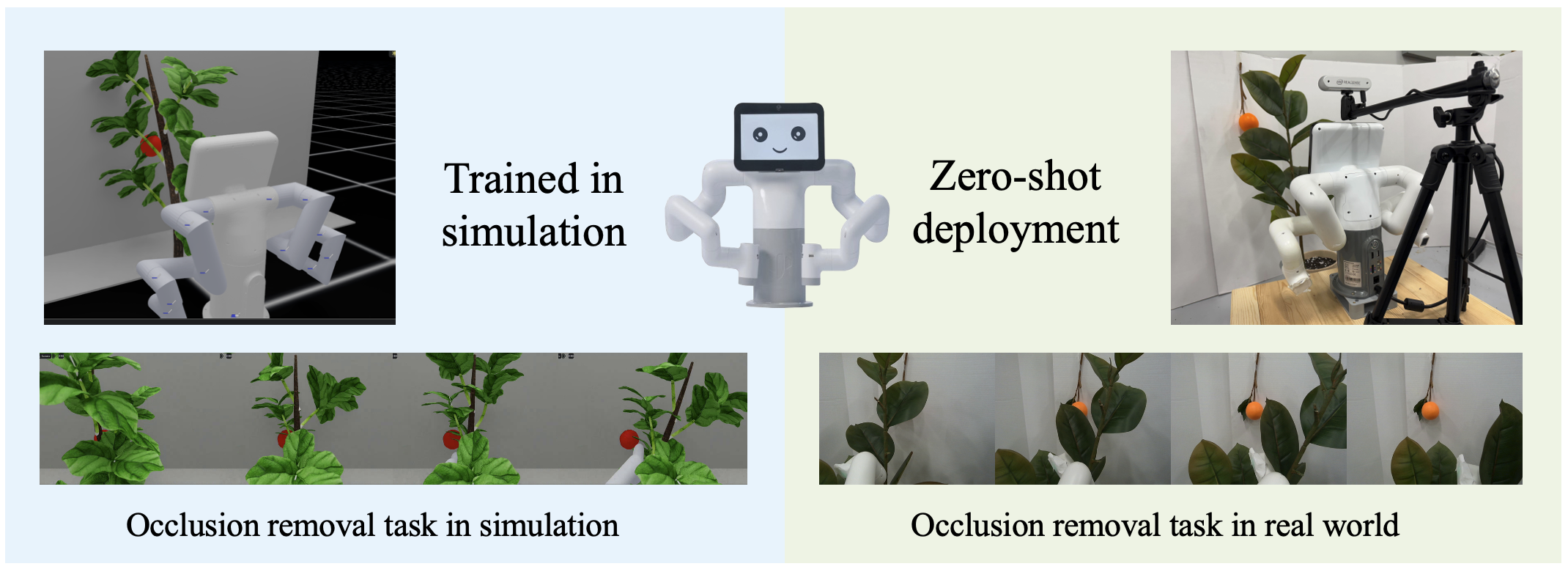}
    \captionof{figure}{\textbf{Top:} We present a sim2real approach for occlusion-aware plant manipulation to reveal fruits in cluttered environments. We perform end-to-end RL training on a generic deformable plant in simulation (left), and then deploy the trained policy in a zero-shot manner on a real stemmed plant (right) using the MyBuddy 280 robot (center). \textbf{Bottom:} Frame-by-frame sequences of occlusion removal in simulation (left) and real-world trials (right), showing that the policy trained on an abstract model generalizes to experimental plants.}
    \label{fig:abstract_figure}
\end{strip}


\begin{abstract}
Autonomous harvesting in the open presents a complex manipulation problem. In most scenarios, an autonomous system has to deal with significant occlusion and require interaction in the presence of large structural uncertainties (every plant is different). Perceptual and modeling uncertainty make design of reliable manipulation controllers for harvesting challenging, resulting in poor performance during deployment. We present a sim2real reinforcement learning (RL) framework for occlusion-aware plant manipulation, where a policy is learned entirely in simulation to reposition stems and leaves to reveal target fruit(s). In our proposed approach, we decouple high-level kinematic planning from low-level compliant control which simplifies the sim2real transfer. This decomposition allows the learned policy to generalize across multiple plants with different stiffness and morphology. In experiments with multiple real-world plant setups, our system achieves up to 86.7\% success in exposing target fruits, demonstrating robustness to occlusion variation and structural uncertainty. 
\end{abstract}

\section{Introduction}

Autonomous harvesting robots must do more than grasping a fruit -- they must first \emph{find} it. In most real world scenarios, fruits are generally hidden deep within dense foliage, partially or completely occluded by leaves, stems, or branches. These occlusions slow down harvesting, increase missed detection rates, and often require manual repositioning before automated grasping can even begin. While recent advances have led to robust perception for visible fruits and reliable manipulation for grasping, the problem of occlusion-resolution for harvesting remains underexplored. Most existing harvesting systems either assume fruits are already visible~\cite{song2025gsnbvgeometrybasedsemanticsawareviewpoint, abouzeid2025pointcloudcompletionapproach, velasquez2024compactroboticgrippertandem}, or adjust the robot itself to get a clearer view of the fruit~\cite{li2025enhancedviewplanningrobotic, jun2021towards, arad2020development, weng2024interactiveperceptiondeformableobject}. However, such strategies fail to scale in real scenarios, where visibility could vary with growth stage, fruit distribution, and foliage density. For robust harvesting, a robot needs to actively manipulate the plant to reveal fruits before grasping --- we refer it as \emph{occlusion-aware fruit discovery}. Addressing this problem is essential for enabling downstream systems to function robustly in real-world conditions.

Plant manipulation, at its core, relies on three tightly coupled ingredients: \emph{perception}, to perceive the environment; \emph{planning}, to determine actions based on perception; and \emph{control}, to execute these actions. Each of these components present unique challenges in field environments, where fruits are embedded within dense, deformable vegetation.

Perception for harvesting is challenging due to significant occlusions: leaves, stems, and branches often hide fruits partially or completely, breaking the assumptions of standard detection pipelines~\cite{chu2023o2rnetoccluderoccludeerelationalnetwork, yao2025safeleafmanipulationaccurate, Kang_2020}. Planning is complicated by cluttered and variable plant geometry~\cite{navone2025autonomousroboticpruningorchards, selvam2025selfsupervisedlearningroboticleaf}. Interaction during manipulation is  difficult due to deformable structure of plants. Flexible stems bend, leaves fold, and branches recoil unpredictably under contact. Designing controllers in such settings is difficult since stiffness, damping, and mass distribution vary across species and even between individuals~\cite{gu2023surveyroboticmanipulationdeformable, doi:10.1177/0278364918779698}. These uncertainties make open-loop strategies and model-based control brittle, especially when occlusions change dynamically during interaction.

Existing research on deformable object manipulation does not fully transfer to agricultural settings. Prior work has largely focused on ropes and chains~\cite{chi2024iterative,nair2017combining,10388462}, fabrics~\cite{ha2022flingbot,huang2022mesh,lin2021softgym,boroushaki2021robotic}, or biological tissues~\cite{10480552}, often under controlled conditions with strong modeling assumptions. Agricultural studies have explored leaf removal~\cite{yao2025safeleafmanipulationaccurate,10160731}, navigating through the branches to reach parts of the plant~\cite{jacobgentle}, sequential exploration to predict newly revealed spaces~\cite{zhang2024pushpastgreenlearning}, or branch repositioning prior to harvesting~\cite{10611327,pmlr-v270-jacob25a, rijal2025forceawarebranchmanipulation}. Other efforts target collision-free grasp planning~\cite{7139558,tafuro2022dpmpdeepprobabilisticmotionplanning}. While valuable, many of these methods rely on explicit plant-part detection or detailed 3D reconstruction~\cite{parhar2018deep,jenkins2017online,3563846}, which are computationally costly and hard to scale to highly deformable, variable foliage.


We believe that a closed-loop learning approach offers a promising alternative. As occlusions evolve during manipulation, classical motion planning methods often struggle~\cite{kwon2025empiricalstudypowerfuture,icarte2020actrememberingstudypartially,liu2022partiallyobservablereinforcementlearning}. Reinforcement learning (RL) adaptively couples perception and action, enabling agents to respond to plant deformation without brittle intermediate models. However, sim-to-real transfer remains challenging~\cite{wagenmaker2024overcoming,ugurlu2022sim,DBLP:journals/corr/abs-2009-13303}, particularly for deformable objects with difficult-to-model mechanics~\cite{ryanl4dc}.

\smallskip
\noindent\textbf{Our Approach:} We propose an end-to-end RL framework for occlusion-aware manipulation of deformable plants with zero-shot sim-to-real transfer (see Fig.~\ref{fig:abstract_figure}). The policy is trained entirely in simulation using a generic plant model built in NVIDIA Isaac Lab~\cite{mittal2025isaaclab}, with deformable dynamics and randomized occlusion patterns. In our simulation setup, the plant is represented using a finite element method (FEM) model rather than simplified kinematic chains or reduced-order approximations. This formulation directly captures bending, twisting, and nonlinear deformation of stems and leaves under contact, without relying on lumped stiffness parameters or geometric heuristics. By avoiding lower-order assumptions, the model preserves the continuous deformation behavior of foliage, providing the RL policy with interaction dynamics that are structurally faithful to real plants~\cite{gu2023surveyroboticmanipulationdeformable, pozzi2018efficient}. This high-fidelity representation serves as a key inductive bias, enabling the learned policy to transfer more reliably to physical systems~\cite{LEGUIZAMO2022193}. This model trains the policy entirely in simulation using PPO~\cite{schulman2017proximal}.

To enable real-world transfer, we combine extensive domain randomization with a compliant low-level controller on the real robot, decoupling high-level planning from uncertain contact dynamics. This design maintains robustness even when real plants differ significantly from the simulated model in stiffness or morphology. In addition to single-target discovery, we also validate the policy in scenarios with multiple fruits, where sequential uncovering is required, further highlighting the generality of the approach.

\begin{table}[!ht]
  \centering
  \small                      
  \renewcommand{\arraystretch}{1.1}
  \begin{tabular}{@{}lcccc@{}} 
    \toprule
    \textbf{Ref.} & \textbf{Task} & \textbf{No Recons.} & \textbf{Visual} & \textbf{RL} \\ 
    \midrule
    Jacob et al. \cite{jacobgentle}      & Reach        & \cmark & \xmark & \cmark \\
    Rijal et al. \cite{rijal2025forceawarebranchmanipulation}      & Branch      & \xmark & \cmark & \xmark \\
    Zhang et al. \cite{zhang2024pushpastgreenlearning}      & Explore      & \cmark & \xmark & \xmark \\
    Yao et al. \cite{yao2025safeleafmanipulationaccurate}        & Leaf       & \xmark & \cmark & \xmark \\
    Jenkins et al. \cite{jenkins2017online}    & Stalk        & \xmark & \cmark & \xmark \\
    \textbf{Ours}         & Adaptive     & \cmark & \cmark & \cmark \\
    \bottomrule
    \end{tabular}
  \caption{Comparison of representative approaches in agricultural or deformable manipulation. ``No Recons.'' indicates no use of explicit 3D reconstruction or geometric modeling; ``Visual'' denotes visual servoing; ``RL'' refers to reinforcement learning. Task labels indicate the primary focus: \emph{Branch} and \emph{Leaf} involve respective manipulation, while \emph{Stalk} refers to its detection.}

    \vspace{2mm}
  \label{tab:comparison_table}
\end{table}

\noindent To the best of our knowledge, this is the first end-to-end RL framework for occlusion-aware fruit discovery that transfers zero-shot to real-world conditions. Table~\ref{tab:comparison_table} provides a comparison of our approach with representative prior works. More specifically, we:
\begin{itemize}
    \item Introduce a task formulation for fruit discovery as a prerequisite to harvesting, targeting plant manipulation without explicit 3D reconstruction,
    \item Propose a decoupled architecture that combines RL-based kinematic planning with a compliant execution layer, improving robustness across plant morphologies,
    \item Demonstrate successful sim-to-real transfer, achieving up to 86.7\% occlusion-resolution success in physical trials (vs. 96.1\% in simulation) using only a generic plant model, and
    \item Highlight policy generalization, showing robustness across individuals of the same morphological class and extending naturally to multiple-fruit scenarios where sequential discovery is required.
\end{itemize}

\begin{figure*}[h]
    \centering
    \includegraphics[width=0.9\linewidth]{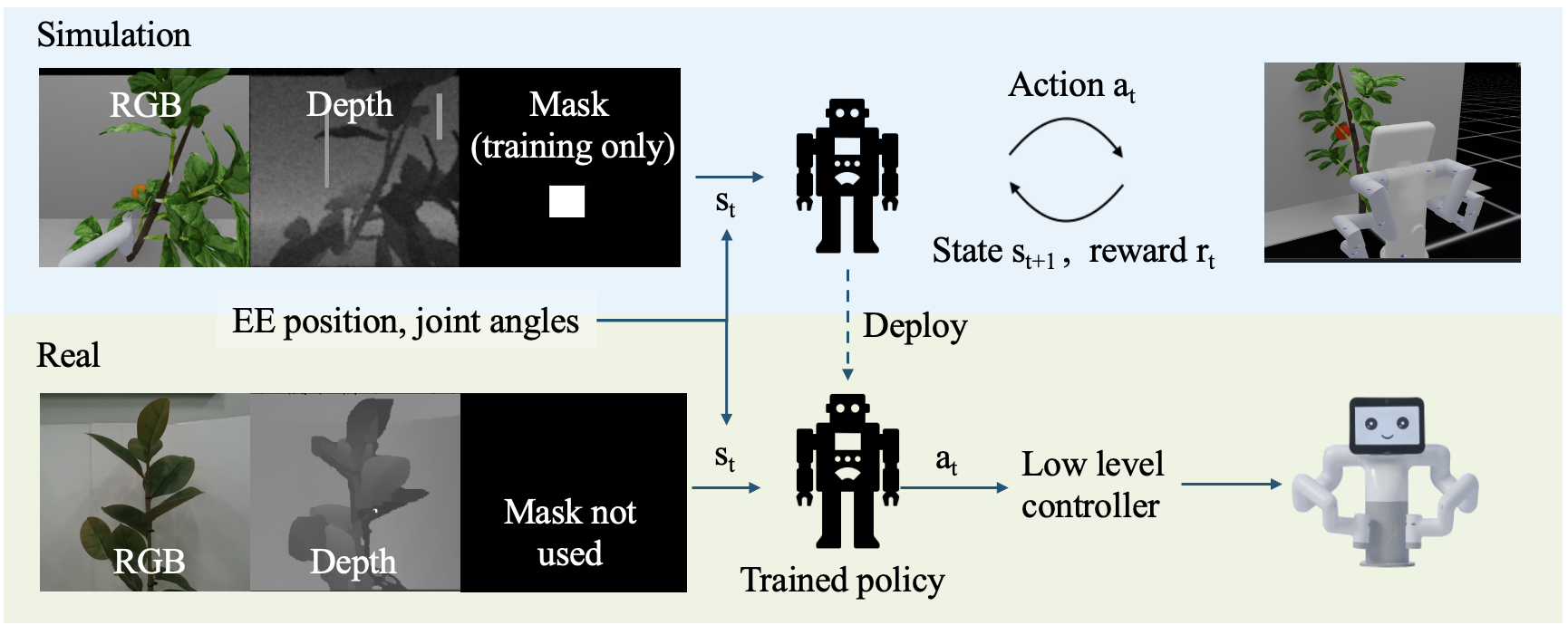}
    \caption{
Overview of the framework. 
\textbf{(a) Sim:} The robot interacts with an abstract plant in Isaac Lab, receiving RGB, depth, and fruit mask inputs (mask used only in training). The RL agent exchanges actions and rewards with the simulator to learn occlusion-aware manipulation. 
\textbf{(b) Real:} The trained policy is deployed on the MyBuddy 280 robot. An RGB-D camera provides observations, and a low-level controller executes high-level commands for safe interaction with real plants.
}
\label{fig:methodology}
\end{figure*}
\section{Methodology}
Our objective is to train a policy that, given visual and proprioceptive inputs, produces a sequence of manipulations to expose a designated fruit hidden by plant foliage. The key challenge lies in bridging the gap between high-level task intent (“expose the fruit”) and low-level contact execution in an environment with uncertain, deformable dynamics.

To address this, we adopt a hierarchical control strategy: \textbf{(1) High-level RL policy:} Operates in configuration space, outputting incremental joint-angle commands that plan how to reduce occlusion over time. \textbf{(2) Low-level controller:} Executes these commands on the real robot, handling contact forces safely and adapting to plant stiffness without requiring the policy to model exact dynamics.


This decoupling enables the policy to learn generalizable strategies in simulation, where physical parameters are randomized, and transfer them to hardware without re-tuning for each plant. The policy leverages structural cues such as the central stem and adapts motion to the target’s visibility state, producing compliant, task-directed interaction.

\subsection{Reinforcement Learning Problem Formulation}
\label{sec:prob_form}
We present the MDP formulation for training the RL policy, including state, action, and reward design as follows:
\begin{itemize}
    \item State: The state $s_t$ at time $t$ is a composite of multi-modal inputs and proprioceptive data. Each state $s_t$ includes an RGBA-D image ($I_{\text{RGBA-D},t}$, RGB image, alpha fruit mask, depth), joint angles for the waist ($J_{b,t}$) and the left arm ($J_{i,t}$, where $i=1\sim5$, representing the different joints), and end-effector position ($EE_{\text{pos},t}$). It can be formally represented as $s_t= [I_{\text{RGBA-D},t}, J_{b,t}, J_{i,t}, EE_{\text{pos},t}]$.

    \item Action: The action $a_t = [\Delta j_{1\text{-}5}, \Delta j_b]$ are delta joint angles for the same joints, clipped to $[-1, 1]$ and scaled at runtime.

    \item Reward: The total reward $r_t$ is a weighted sum of:\\
    (a) Occlusion reward, $r_{occ}$ = (1 - \text{occluded / fruit pixels})  \\
    (b) Full visibility reward, $r_{\text{fv}} = 3$, upon satisfying the success criterion for 1 timestep (Section~\ref{sec: evaluationmetrices}).  \\
    (c) Sustained visibility, $r_{sus} = 20$ if full visibility is maintained $\geq 10$ steps; 0 otherwise.  \\
    (d) Post-visibility penalty, $r_{pv} = \text{action magnitude}$ after full visibility, penalizing post-visibility motion.\\
    (e) Self-collision penalty, $r_{sc} = -5$ if self collision; 0 otherwise.
\end{itemize}
Together, with the selected weights for each term, the reward function becomes:
\begin{align}
r_i = &\; 10.0 \cdot \left(1 - \frac{P_{\text{occ}, i}}{40 \times 40}\right)
+ 3.0 \cdot \mathds{1}_{[P_{\text{occ}, i} \leq 160]} \nonumber \\
&+ 20.0 \cdot \mathds{1}_{\text{sus}, i} 
- 0.06 \cdot \|\mathbf{a}_i\|_2 \cdot \mathds{1}_{[P_{\text{occ}, i} \leq 160]} \nonumber\\
&- 5.0 \cdot \mathds{1}_{sc, i}
\end{align}

\subsection{Simulation Environment Setup}
The simulation environment, as shown in top right of Fig.~\ref{fig:methodology}, is developed using Isaac Lab from NVIDIA Omniverse, ensuring high-fidelity replication of the physical setup. The MyBuddy 280 robot is imported as an URDF model derived from SolidWorks, with joint positions controlled to mirror the behavior of the physical robot closely. Artificial plants are modeled as OBJ meshes converted into deformable bodies using Isaac Lab’s API. Primitive shapes like spheres and cubes were randomly placed behind the plant to mimic the variable fruit location. Domain randomization is applied to improve robustness, including 360° variations in plant orientation and perturbed lighting conditions. The PhysX engine operates at a 1/60 s timestep with eight solver iterations. A white wall is placed behind the fruits to minimize background distractions. All simulations are conducted in Isaac Lab 1.4.0 with an NVIDIA GeForce RTX 4090 GPU.

Although simplified, the simulated plant captures essential structural features: a central stem, occluding leaves, and a fruit target. This design encodes an inductive bias toward realistic manipulation strategies, encouraging the policy to discover affordances such as sliding along the stem or pushing leaves aside, which transfer naturally to real plants. Randomization of appearance, stiffness, and fruit placement prevents overfitting and ensures generalization.

\subsection{Evaluation Metrics}
\label{sec: evaluationmetrices}
To assess the RL agent's performance, we define the following evaluation metrics:
\begin{itemize}
    \item Success rate: the percentage of trials in which at least 90\% of the fruit is evident for more than 10 timesteps.

    \item Steps to goal: The number of steps required to achieve the visibility objective.

    \item Generalization: The percentage of successful trials across various configurations and environmental setups.
\end{itemize}

\subsection{Training and Evaluation of the RL Agent}
\label{subsec: Training and Evaluation}

\begin{figure*}[!t]
    \centering
    \includegraphics[width=0.85\textwidth]{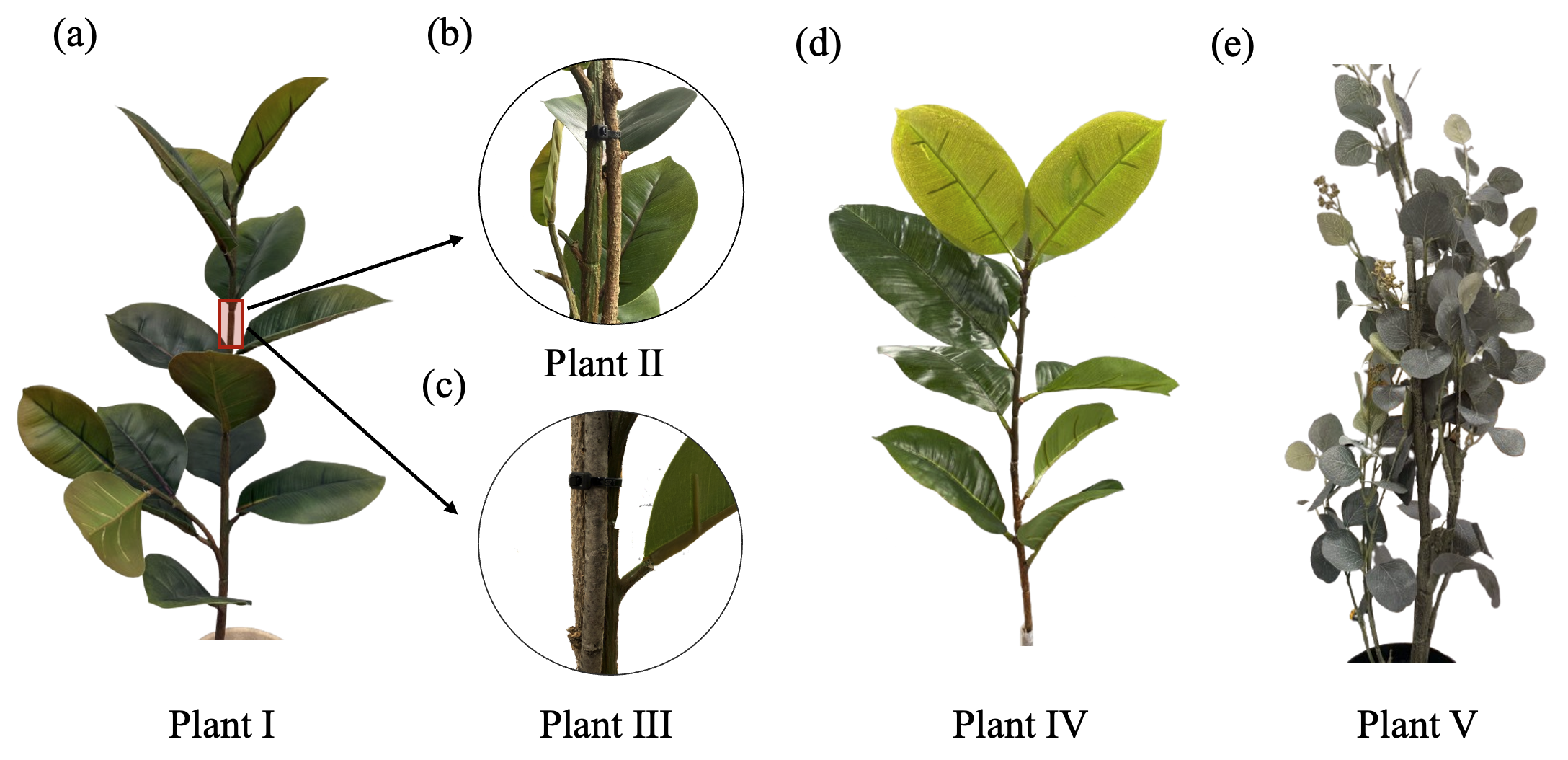}
    \caption{Plant used for real-world experiments. \textbf{(a)} and \textbf{(d)} show plants with a clearly visible central stem and broad foliage, similar to the simulated morphology. \textbf{(b)} and \textbf{(c)} show reinforced versions of \textbf{(a)}: \textbf{(b)} is the doubled reinforced version of \textbf{(a)}, and \textbf{(c)} is the tripple reinforced version of \textbf{(a)}. The zoom-in highlights the reinforcement detail. \textbf{(e)} depicts a dense, bushy plant lacking a distinct central stem, presenting a more severe occlusion challenge. Together, these variations enable testing of the policy's robustness across a range of plant geometry and mechanical properties. All plants are artificial.}
    \label{fig:plant_comparison}
\end{figure*}
\paragraph{Training} We train the policy using the PPO ~\cite{schulman2017proximal} implementation from skrl~\cite{serrano2023skrl} in simulation (please see Appendix for training configurations, the link to appendix is present in footer of page 1), enabling agent to learn continuous control of the robot’s 6-DOF arm. Each episode begins with a randomized robot configuration and fruit location, requiring the agent to adapt to varying views and occlusion. To promote generalization, we apply domain randomization over physical properties (torque), visual conditions (lighting, textures), and sensor noise. These variations prevent overfitting to a specific environment and bolster robustness to real-world deployment.

To facilitate efficient learning, we provide the agent with privileged information during training: a ground-truth binary mask highlighting the fruit's location. This auxiliary input guides the agent toward identifying the fruit, especially when it is heavily occluded, and accelerates the acquisition of effective behaviors. Importantly, the mask is used only during training; at evaluation and deployment, the policy relies solely on sensory observations (see Section~\ref{subsec:ablations}).

After approximately 50 million simulation steps, the agent converges to a policy that reliably finds and exposes fruit by strategically interacting with the plant: locate the stem and fruit, move the end-effector along the stem, and push aside occluding structure until the fruit becomes visible. This behavior, shaped by training in diverse randomized environments, generalizes well to real plants with similar stems and occlusion patterns.

\paragraph{Evaluation} During post-training evaluation in simulation, 20,000 randomized episodes mirroring training conditions, the RL agent achieved an overall success rate of \textbf{96.09\%} (see Appendix's Fig. 1 for the plots). In most trials, the agent revealed the fruit with minimal occlusion, demonstrating the robustness and effectiveness of the learned policy. The agent also exhibited efficient behavior, requiring as few as 200 control steps in the easiest scenarios and up to 600 in the most challenging ones (mean episode length: $\approx\!500$ steps). This range indicates adaptability and consistency: even from unfavorable starting positions or highly occluded setups, the policy reliably finds a sequence of actions to achieve visibility.

\subsection{Deployment Framework}
\label{subsec:deploy}

The deployment framework (Fig.~\ref{fig:methodology}) integrates the trained RL controller with a low-level servo position controller. The end-effector pose and joint angles are published to ROS 2 and subscribed to by the control PC. Simultaneously, an RGB-D image is captured from the sensor. These observations are fed into the policy, which generates an action. The action, converted to target joint angles, is then published and received by the robot. At this stage, only kinematic control is used. The servo controller tracks the target position in a controlled manner without exerting excessive force on the plant. This process runs in a continuous loop, ensuring safe and effective control.

\begin{figure*}[!t]
    \centering
    \includegraphics[width=0.95\textwidth]{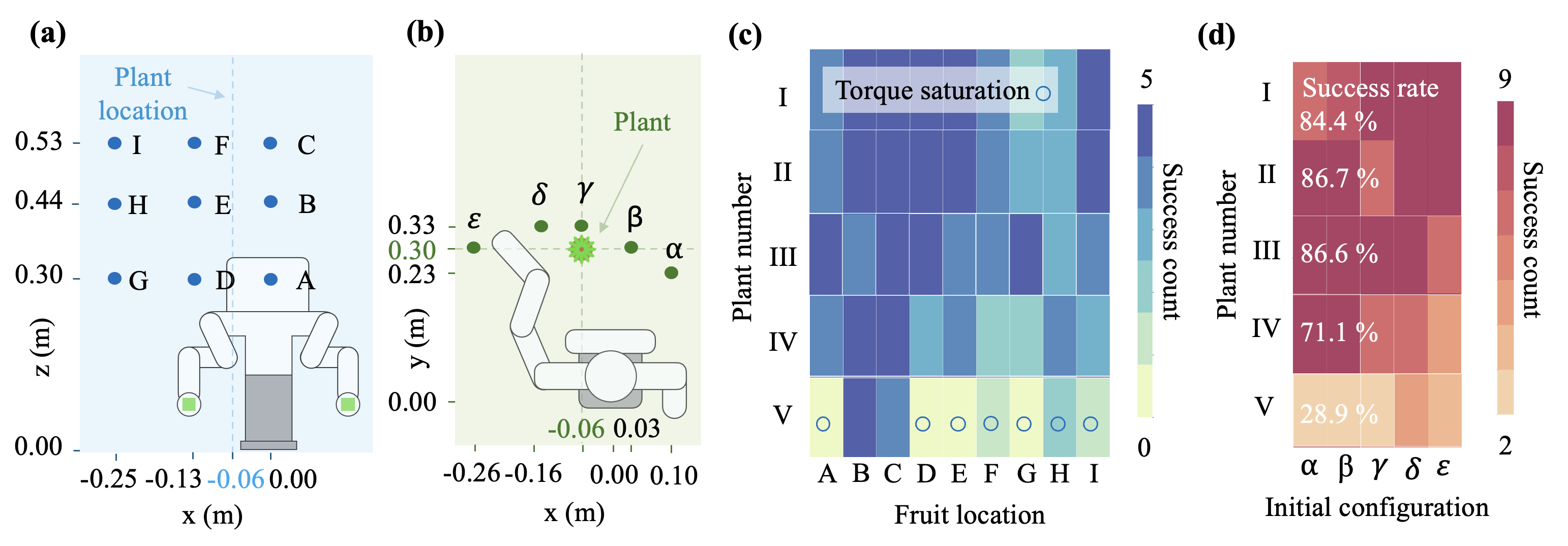}
    \caption{\textbf{(a)} Experimental setup showing predefined fruit locations relative to the robot and the plant, viewed from the rear of the robot (x-z plane). \textbf{(b)} Top-down view (x-y plane) of the five different initial configurations, presented in end-effector positions. \textbf{(c)} Heatmap of successful trials across all plants  (I-V) and fruit locations (A-I), where success count represents the number of successful attempts out of five trials for each initial configuration. \textbf{(d)} Heatmap of successful trials across all plant (I-V) and initial configurations ($\alpha$-$\varepsilon$), where success count represents the number of successful attempts out of nine trials for each fruit location The percentage numbers at the lower left are the success rates across each plant.}
    \label{fig:exp_results}
\end{figure*}

\section{Experiments}
\label{subsec:experiments}
After training exclusively in simulation, we deployed the policy zero‑shot on a physical setup (Fig.~\ref{fig:abstract_figure} top) consisting of a 6‑DoF MyBuddy arm, a RealSense camera~\cite{keselman2017intel}, and a selection of potted, off‑the‑shelf artificial plants. Camera height, distance, and tilt as well as the plant’s base pose-were fixed to mirror the simulation.

To evaluate generalization, we conducted experiments across a structured grid of initial conditions. Five plant instances, presented in Fig.~\ref{fig:plant_comparison}, were considered: two morphologically similar  (Plant I and IV), two structurally reinforced variant of one of the former two with increased stiffness (Plant II and Plant III), and a visually distinct, bushy plant with the stiffest stem (Plant V). Each plant exhibits unique stiffness characteristics, enabling evaluation of the policy's robustness across diverse physical and visual conditions. For each plant, the fruit was placed at nine discrete locations, arranged in a \(3\times3\) lattice spanning left/center/right and low/mid/high positions, 20~cm behind the plant (Fig.~\ref{fig:exp_results}(a)). The robotic arm was initialized from five pre-selected, self-collision-free joint configurations (Fig.~\ref{fig:exp_results}(b)), resulting in \(5 \times 9 \times 5 = 225\) unique trials. In each trial, the arm was reset to its designated pose, the fruit positioned at the target grid cell, and the trained policy was executed for up to 600 control steps at a frequency of 1~Hz, under low-level compliance control. A trial was considered successful if the robot achieved at least 90\% fruit visibility for five consecutive frames. The number of control steps required to achieve success was also recorded.

\section{Results}
\label{sec:results}


This section presents results from deploying our learned policy on the real setup, focusing on observed behaviors, success rates, and failure modes. We first analyze generalization across plant morphologies, followed by evaluation in sequential multi-fruit scenarios and ablations on privileged information and simulation fidelity. Finally, we discuss key failure cases and limitations.

\begin{figure*}
    \centering
    \includegraphics[width=0.95\linewidth]{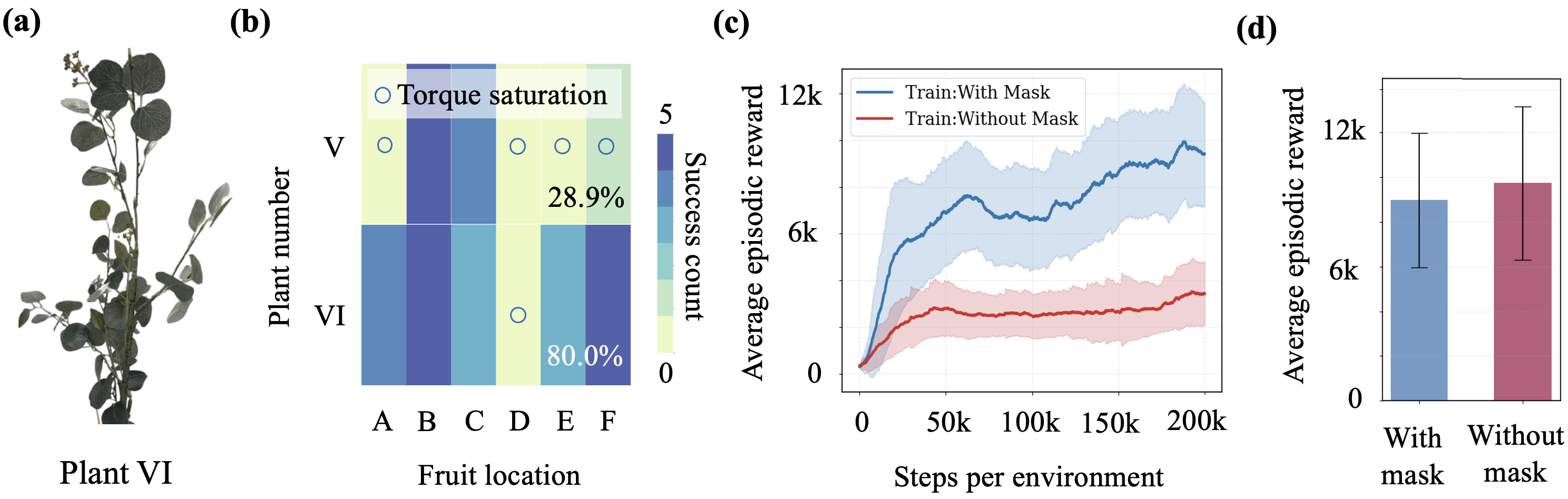}
    \caption{\textbf{(a)} The modified version of Plant~IV: mechanically more compliant while preserving its overall structure and visual complexity \textbf{(b)} Heatmap of successful trials between Plant~IV and V and fruit locations (A-F), where success count represents the number of successful attempts out of five trials for each initial configuration. \textbf{(c)} Training performance with and without access to the ground-truth fruit mask. The mask accelerates learning and improves final returns \textbf{(d)} Evaluation results when deploying the policy trained with the mask: performance remains consistent even when the mask is removed at test time
    }
    \label{fig:ablation}
\end{figure*}

\subsection{Policy Behavior and Generalization}

In most trials, the robot exhibited purposeful and consistent behavior closely resembling what was observed in simulation. Based on our observations, the policy inferred likely fruit locations and anticipated occlusions. It then guided the end-effector toward occluded regions using compliant, stem-leveraging motions: the robot often maneuvered around the central stem, pushing it aside to expose hidden fruit. Once visibility exceeded the success threshold, the robot typically ceased further movement, demonstrating task-directed interaction.

As summarized in Fig.~\ref{fig:exp_results}(c) and (d), the policy achieved high success rates on Plant~I (84.4\%), Plant~II (86.7\%), and Plant~III (86.6\%), Plant~IV (71.1\%) confirming strong generalization across different mechanical properties. Performance dropped on Plant~V (28.9\%), a bushy plant with no visible central stem and increased stiffness. The two key contributing factors for this drop are: (1) perception challenges due to the different morphology of the testing plant, and (2) increased stem stiffness, the dominant limitation, since the required motion exceeded the robot’s torque capacity, leading to torque saturation. Our additional experiments corroborate this attribution (see Modified stiffness test in Section IV-D).

\subsection{Multiple Fruit Generalization} 
Our policy is trained primarily in a single-fruit setup. Nevertheless, it could generalizes to multiple targets without architectural changes by following a \emph{sequential exposure} protocol: once one fruit is uncovered, it is removed from the scene, and the same policy continues searching for the remaining fruit(s). This removal reflects a practical harvesting scenario in which one arm exposes the fruit while another arm can perform the picking task. Importantly, the control loop runs continuously, we do not reset the robot or environment after the first fruit is found. In scenarios where the first fruit is not to be harvested, e.g., unripe, though not implemented in our experiments, we could apply an ignore mask in the observation stream to prevent re-targeting while proceeding to the next fruit. The testing procedures are:

\textbf{Setup:} Two fruits were placed behind foliage in Plant~I across two initial arm configurations ($\alpha$ and $\epsilon$). The observation/action spaces were unchanged. Once a fruit reached the success criterion ($\geq 90\%$ visible for 10 frames), it was physically removed and the same policy continued operating without resetting the robot or environment.

\textbf{Results:} Across 18 trials, both fruits were exposed in 16 cases (88.9\%). Failures occurred only in heavily occluded placements under the $\epsilon$ initialization, where torque limits prevented full exposure. On average, the first fruit required 40--400 steps, while the second required only 50--150 steps, due to the reduced occlusion.

\begin{figure}[h]
  \centering
  \includegraphics[width=\linewidth]{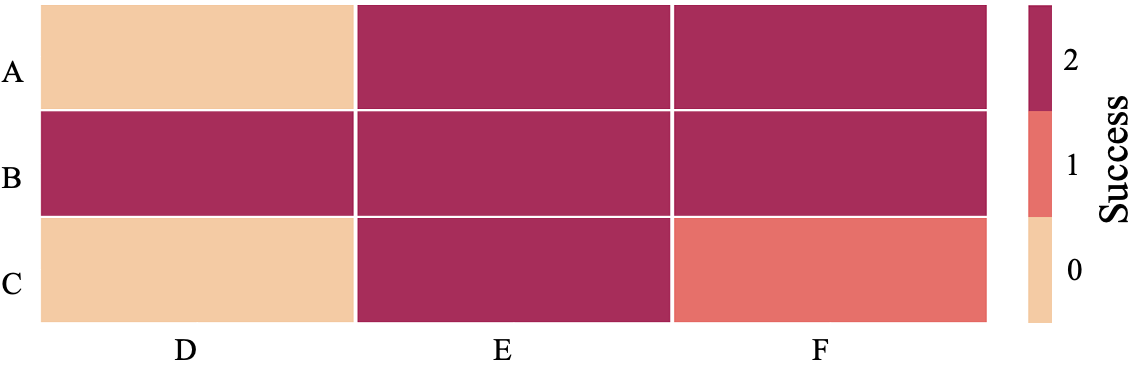}
  \captionsetup{skip=0pt}
\caption{
Multi-fruit exposure success across different fruit location pairs. Rows indicate the first fruit location (A--C) and columns indicate the second (D--F). Each cell shows the number of fruits successfully uncovered (0--2) in sequential trials. The policy achieves high success for most pairs, with failures occurring primarily when both fruits are heavily occluded (pairs A--D and C--D).
}
\vspace{3mm}
\label{fig:multifruit_bar}
\end{figure}

In $18$ two-fruit trials on Plant I, the policy successfully exposed both fruits in 16 cases (88.9\%). Failures occurred only when both fruits were placed in heavily occluded positions (pairs D--A and D--C) under the \emph{$\epsilon$} initialization, where torque saturation prevented full exposure. In all other fruit pairs, success was achieved across both initial configurations. On average, the first fruit required 300--400 steps to expose, while the second fruit was exposed within 50--150 steps, since removal of the first target reduced occlusion for the next. These results confirm that the single-fruit policy extends naturally to sequential multi-fruit exposure without retraining. In practice, we observe the same stem-leveraging behavior as in the single-target case: the end-effector follows the stem, displaces occluding leaves, and halts once visibility is achieved. Removing the first fruit often relaxes local occlusions, reducing the steps needed for the second target. Failures typically arise from (1) re-occlusion due to plant relaxation after removal and (2) torque saturation on stiff stems, mirroring single-target limits. Please see Fig.~\ref{fig:multifruit_bar} for full results of the multi-fruit tests. 

\subsection{Ablation Studies}
\label{subsec:ablations}

To better understand the critical factors behind successful training and sim-to-real transfer, we conducted two ablations:

\textbf{Privileged fruit mask during training:} During training, we supplied a ground-truth fruit mask as an additional input channel. As shown in Fig.~\ref{fig:ablation}(c) and (d), policies trained with this privileged signal converged faster and reached higher returns. Importantly, removing the mask at test time had negligible effect, confirming that the mask accelerates learning but the final behavior relies only on RGB-D and proprioceptive inputs. This validates our training-time design choice.

\textbf{Simulation fidelity:} We trained two policies: one in a high-fidelity simulator with realistic textures and lighting, and another in a reduced-fidelity variant with flat textures and simplified illumination. The high-fidelity policy transferred successfully, achieving up to 86.7\% real-world success. In contrast, the low-fidelity policy exhibited erratic behavior and failed to generalize, preventing real-world deployment. This demonstrates that while precise digital twins are unnecessary, sufficient visual realism in simulation is critical for effective sim-to-real transfer.

\subsection{Failure Modes}
\textbf{Perception artifacts:} We observed reduced performance on Plant~II despite its similarity to Plant~I. RealSense depth images occasionally produced ghosted or duplicated stems, leading to ambiguous depth cues and misinterpretation by the policy. These cases underscore the importance of sensor fidelity alongside policy robustness. Representative RGB-D frames are shown in Appendix's Figs. 7 - 10.

\textbf{Torque saturation:} The dominant failure mode on Plant~IV was mechanical: the robot often saturated its torque limits, preventing execution of the commanded trajectory. Even in failed trials, partial exposure of 70--80\% was achieved, just below the 90\% success threshold. 

\textbf{Modified stiffness test:} To test whether failures were due to mechanical resistance rather than policy limitations, we created Plant~VI by softening Plant~IV while preserving its visual complexity (Fig.~\ref{fig:ablation}(a)). On Plant~VI, the policy achieved 80\% success across valid fruit positions, comparable to Plants~I–III and Plant~V. Including a torque-limited fruit location (D) reduced the overall success to 67\%. These findings confirm that the learned policy generalizes to visually distinct plants as long as actuation demands remain within hardware limits.

\subsection{Summary of Findings}

Overall, our results demonstrate that:
\begin{itemize}
    \item The policy generalizes well across stemmed plants with varying stiffness, achieving up to 86.7\% success in real trials.
    \item Failures arise primarily from mechanical resistance exceeding actuation limits and from sensor artifacts in depth perception.
    \item Privileged fruit masks accelerate training but are unnecessary at test time.
    \item High-fidelity finite element but abstract plant models suffice for sim-to-real transfer; exact digital twins are not required.
\end{itemize}

These results suggest that a simulated plant with a central stem and occluding leaves offers enough structural cues to support robust real-world deployment.

\section{Conclusion and Future Work}
\label{sec:conclusion}
We presented an end-to-end RL framework for occlusion-aware fruit discovery, trained entirely in simulation and deployed zero-shot to real plants. The approach combines a high-level kinematic policy with a compliant low-level controller, allowing the robot to uncover occluded fruits safely and adaptively while leveraging the central stem as a structural cue. Robust sim-to-real transfer is achieved through extensive domain randomization and the use of privileged signals during training, without requiring fine-tuning or very accurate digital twins. A key factor in this transfer is our high-fidelity finite element (FEM) plant model, which captures realistic contact dynamics without reduced-order approximations. By exposing the policy to nonlinear deformations of stems and leaves, the FEM model enables strategies that generalize across plant morphologies and multi-fruit scenarios, providing a critical bridge between abstract simulation and the complex behavior of real foliage.

In real-world trials, the learned policy achieved up to 86.7\% success on stemmed plants, closely matching simulation performance (96.1\%). Notably, our simulation relied on a generic morphological model rather than plant-specific reconstruction, demonstrating that abstract but structurally representative models are sufficient for transfer. These results mark a step toward scalable, perception-driven agricultural robots capable of reliable operation in cluttered, deformable environments.

Looking forward, we aim to expand the simulation framework to multi-branch canopies and fruit clusters, providing richer training scenarios. Beyond occlusion resolution, a key next step is adding a downstream fruit-picking task using the robot’s second arm. Training in more realistic scenarios; such as fruits located deep inside foliage where both arms may need to coordinate to first clear occlusions and then grasp the target; represents an exciting future direction toward fully autonomous harvesting.

\section{Limitations}
\label{sec:limitations}

Our approach shows strong performance, but several limitations remain:

(a) Simplified simulation: Training relied on a single-stem, single-fruit abstraction. While adequate for simple plants, it does not capture complex morphologies like multi-branch canopies or fruit clusters. We also assumed static fruit state and a fixed overhead camera. Extending the simulator to relax these assumptions is a natural next step.

(b) Stem dependence: The policy assumes access to the central stem. In dense or constrained foliage this may not hold. Incorporating tactile feedback, contact-aware planning, or alternative strategies could improve robustness.

(c) Sim-to-real gap: Depth artifacts, lighting variation, and unmodeled dynamics (e.g., wind) still affect performance. Real-to-sim adaptation and more invariant visual encoders are promising remedies.

(d) Hardware limits: Failures on stiff plants highlight torque saturation. Stronger or variable-compliance actuators could expand applicability.

In summary, while precise digital twins are not required for effective sim-to-real transfer, addressing these limitations will be critical for scaling to field-deployable systems. Broader plant morphologies, richer occlusion scenarios, and enhanced perception robustness represent promising directions for enabling autonomous harvesters to operate reliably in diverse agricultural settings.

\section*{Acknowledgment}
This work is jointly funded by the NSF–USDA COALESCE program (NSF award \#1954556; USDA award \#2021-6702134418)

\bibliographystyle{ieeetr}
\bibliography{references}

\end{document}